\documentclass[a4paper]{article}

\usepackage{INTERSPEECH2021}
\usepackage{cite}

\title{Towards Automatic Speech to Sign Language Generation}
\name{Parul Kapoor$^1$, Rudrabha Mukhopadhyay$^2$, Sindhu B Hegde$^2$, Vinay Namboodiri$^{1,3}$, C V Jawahar$^2$  }
\address{
  $^1$Department of Computer Science and Engineering, IIT Kanpur\\
  $^2$Center for Visual Information Technology, IIIT Hyderabad\\
  $^3$Department of Computer Science, University of Bath}
\email{kparul@iitk.ac.in, \{radrabha.m, sindhu.hegde\}@research.iiit.ac.in, vpn22@bath.ac.uk, jawahar@iiit.ac.in}

\begin{document}

\maketitle  
\begin{abstract}
We aim to solve the highly challenging task of generating continuous sign language videos solely from speech segments for the first time. Recent efforts in this space have focused on generating such videos from human-annotated text transcripts without considering other modalities. However, replacing speech with sign language proves to be a practical solution while communicating with people suffering from hearing loss. Therefore, we eliminate the need of using text as input and design techniques that work for more natural, continuous, freely uttered speech covering an extensive vocabulary. Since the current datasets are inadequate for generating sign language directly from speech, we collect and release the first Indian sign language dataset comprising speech-level annotations, text transcripts, and the corresponding sign-language videos. Next, we propose a multi-tasking transformer network trained to generate signer's poses from speech segments. With speech-to-text as an auxiliary task and an additional cross-modal discriminator, our model learns to generate  continuous sign pose sequences in an end-to-end manner. Extensive experiments and comparisons with other baselines demonstrate the effectiveness of our approach. We also conduct additional ablation studies to analyze the effect of different modules of our network. A demo video containing several results is attached to the supplementary material.
\end{abstract}


\section{Introduction}

According to a World Health Organization report~\cite{pmid24839326}, over $466$ million people worldwide, or roughly $5\%$ of the world's population, suffer from hearing loss. Sign language is often the primary means of communication used by people with hearing disabilities. 
Sign language consists of manual communication features including hand gestures, hand-shape, location, movement, orientation. It includes non-manual gestures like eye gaze, eyebrows, and mouth movement. 

Sign-language is popularly represented in human-readable form using glosses and text transcripts. However, annotating at the gloss level is a tedious task and limits datasets to a smaller size. While efforts~\cite{10.1007/978-3-030-58621-8_40, 9093516} have been focused on using text as the input modality, this is inherently limiting. Generally, sign-language replaces speech as a communication medium for people suffering from hearing loss. Thus, apart from the content, the speaker's emotions and other attributes are also communicated via sign language. This additional stream of information is completely lost when converting text to corresponding sign language videos. Text transcripts are also annotated at the sentence level, limiting the amount of higher-level context information provided for better continuous sign language generation. Finally, most applications like interviews and news reporting require direct translation from continuous speech to sign language that cannot be segmented into sentences.

In this work, we focus on overcoming the current setup's shortcomings and directly translate natural speech to the corresponding sign language poses for the first time. This allows us to generate improved sign-language gestures by inherently transferring content and style from the speech. To generate sign language videos from speech, we collect and release the first large-scale dataset tailor-made for our task. The dataset contains videos of a signer signing in Indian Sign Language standards~\cite{sinha2018indian} accompanied by corresponding speech segments. Each speech segment also contains the corresponding text transcripts. We then train a multi-task transformer network on the collected data coupled with a cross-modal discriminator to generate intelligible sign-language videos from the given speech segment.
The proposed dataset and code will be released for future research at https://kapoorparul.github.io/S2SL/

\begin{figure}[t]
  \centering
  \includegraphics[width=\linewidth]{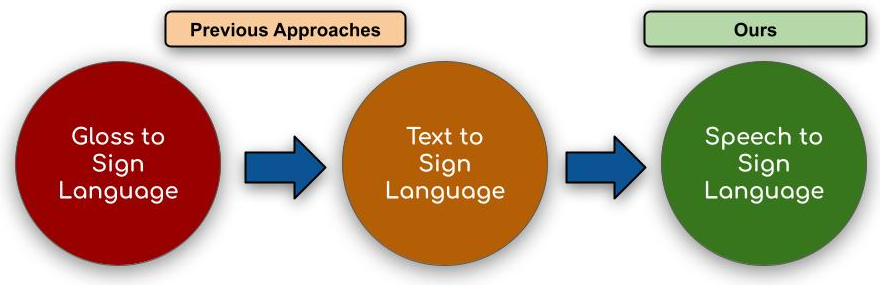}
  \caption{Previous approaches have only attempted to generate sign-language from text transcripts, we focus on directly converting speech segments into sign-language. Our work opens up several assistive technology applications and can help effectively communicate with people suffering from hearing loss. }
  \label{fig:banner}
  \vspace{-10pt}
\end{figure}

\section{Text to Sign Language Generation}
Sign-language generation from text has always been an essential part of assistive technologies. Initial works~\cite{tessa, Glauert2006VANESSAA, wyvill1986animating} for this problem depended on classical grammar-based approaches to stitch together signs for isolated words to achieve continuous sign language production. Like in many other cases, deep-learning has had a massive impact on modern techniques. Researchers considered sign-language as another ``language" and modeled converting the text in language $A$ to ``sign-language" as a Neural Machine Translation (NMT) problem. This approach was used in~\cite{surrey848809,Stoll2020} with varying degrees of success. Zelinka et al.~\cite{9093516} proposed generating Czech sign language from words rather than glosses but they restrict the length of signs to $7$ frames for each word and work with a very limited vocabulary of $598$ Czech lemmas. Recent methods~\cite{10.1007/978-3-030-58621-8_40, surrey858417} use transformer networks to learn long-term dependencies between the input and output modalities. Most of these works provide text tokens as input and generate a sequence of human poses as the signed output. 

Even though these works show decent improvements in the generation of sign-language from text, there has been no push to include speech as the input. Some previous works~\cite{10.1016/j.specom.2008.02.001,Paulson2014AnAS,inproceedings} propose speech to sign language but rely on intermediate automatic speech recognition step and finally use rule based approaches to convert text to sign language videos for a very restrictive vocabulary. Despite bringing an additional stream of information while communicating, speech as a modality brings a lot of challenges to the current setting. For example, unlike text, speech cannot be segmented into discrete units like words or glosses. The lack of datasets containing speech-level annotations also complicates the task. To help solve this problem, we collect a dataset amenable to our task. We then propose a multi-task transformer-based architecture for learning high-level contextual information between speech and the signer's poses. We also use a cross-modal discriminator to classify whether the generated poses match the input speech. In Section~\ref{s2sl}, we discuss our dataset collection procedure and the proposed network in detail. 
\section{Speech To Sign Language Generation}
\label{s2sl}
In this work, we make the first-ever attempt to generate sign-language directly from the speech input by proposing a fully end-to-end model. Formally, given a speech input sequence $S$ as $\{ s_1,s_2,..,s_n \}$, our aim is to design a model that will generate a human pose key point sequence $P$ as $\{p_1,p_1,..,p_m\} $ corresponding to the sign language. Here $p_i$ corresponds to the pose key points for the $i^{th}$ frame in the sign language sequence. For each input speech segment $S$, the corresponding text is also transcribed as $T=\{t_1,t_2,..,t_Q\}$. Due to the unavailability of datasets containing speech level annotation, we curate the first continuous sign language dataset adhering to requirements. Our sign language dataset is based on Indian Sign Language (ISL) standards as it is curated from Indian news reports that a signer annotates in real-time.

\subsection{Indian Sign Language Dataset}

\begin{figure}[t]
  \centering
  \includegraphics[width=\linewidth]{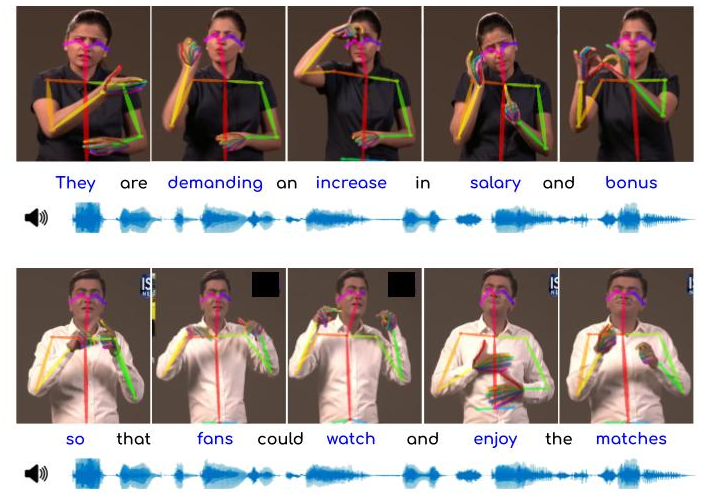}
  \caption{We introduce a new, large-scale Indian Sign language (ISL) dataset containing continuous sign language videos from $5$ signers along with the corresponding text and speech transcripts. Our dataset covers a large number of topics and has a vocabulary of around $10$K English words.
  }
  \label{fig:dataset}
  \vspace{-10pt}
\end{figure}

Over time, several datasets have been proposed for American~\cite{Zahedi06continuoussign}, German~\cite{8578910}, Korean~\cite{app9132683}, and Chinese~\cite{Huang2018VideobasedSL} sign languages. Since these datasets do not contain the required speech annotations necessary for our task, we collect and release a large scale corpus containing all three modalities: (i) continuous sign-language videos, (ii) text and (iii) speech. Our dataset is based on ISL standards~\cite{sinha2018indian} and is far more extensive than the current ISL datasets~\cite{10.1007/978-3-642-12214-9_18,10.1145/3394171.3413528}. 


Our proposed corpus spans a vast vocabulary of $\approx 10$K English words. We have extracted a total of $9092$ high-quality video recordings of the news for the hearing impaired available freely on the internet. Specifically, these videos consist of front-facing professional signers signing the spoken sentences that a hidden speaker says, as shown in Figure~\ref{fig:dataset}. Our data consists of over $18$ hours of videos across $5$ professional signers covering various topics, including current affairs, sports, and world news. A comparison of our dataset with the existing datasets is presented in Table~\ref{table:dataset}. 
As seen from the table, German sign language dataset Phoenix 14T~\cite{8578910} is significantly less challenging in comparison to our dataset due to its limited vocabulary of $2887$ words.
\begin{table}[ht]
    \centering
    \footnotesize
    \setlength{\tabcolsep}{1.2pt}
    \caption{Comparison of our dataset with several other publicly available datasets containing sign language information. Ours is the first continuous Indian sign dataset to contain natural speech annotations.
    }
    \begin{tabular}{lccccccc}
    \hline

    Dataset & lang & vocab & $\#$hours & $\#$signers & continuous & speech \\
    
    \hline 
    KETI~\cite{app9132683} & KSL &  524 &  20 & 14 & \checkmark & $\times$ \\
    Boston104~\cite{Zahedi06continuoussign} & ASL & 104 & 0.7 & 3 & \checkmark & $\times$  \\
    Phoenix 14T~\cite{8578910} & GSL & 3k & 11 & 9 & \checkmark & $\times$\\
    \hline
    INCLUDE~\cite{10.1145/3394171.3413528} & ISL & 263 & - & 7 & $\times$ & $\times$ \\
    \textbf{Ours} & ISL & 10k & 18 & 5  & \checkmark &  \checkmark\\
    \hline
    \end{tabular}
    \vspace{-10pt}
    \label{table:dataset}
\end{table}

\begin{figure}[t]
  \centering
  \includegraphics[width=\linewidth]{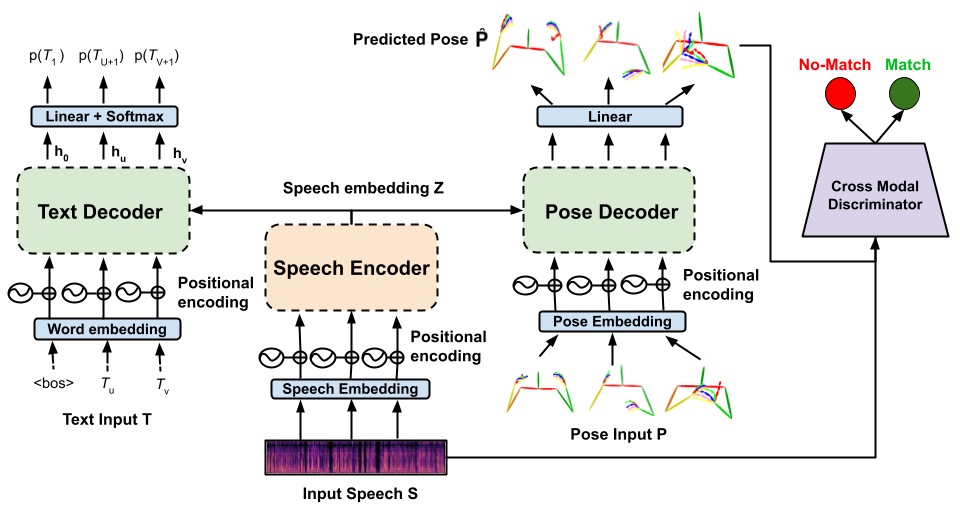}
  \caption{Multi-task transformer architecture which generates sign language pose sequences along with the secondary task of generating the text translations for the given speech input.}
  \label{fig:transformer}
  \vspace{-10pt}
\end{figure}

\subsection{Multi-tasking Transformer Network}
Generating sign language poses solely from speech is a highly challenging task. We propose a multi-tasking transformer network coupled with cross-modal discriminator to generate poses directly from speech segments. We describe the different modules of our network in this section.  

The backbone of our proposed model is a modified transformer architecture introduced in~\cite{10.5555/3295222.3295349}. In most natural language processing tasks, both the input and the target domain consist of discrete vocabulary. On the contrary, for our task, both input and target belong to the continuous space. To assist our primary goal of generating sign language poses from speech, we introduce an auxiliary task of recognizing the input speech and train jointly. Our transformer architecture consists of three major components: (i) a joint speech encoder, (ii) a pose decoder and (iii) a text decoder decoding the complementary modalities. Finally, we also employ a cross-modal matching network as a discriminator to help the transformer learn the high-quality translations from speech to sign language. A pictorial representation of our architecture is given in Figure~\ref{fig:transformer}. 

\subsubsection{Generating Poses from Videos}
We use OpenPose~\cite{8765346} to represent the body poses using the human pose joint keypoints for the signers. We consider $50$ human upper body key points including hands and fingers. To minimize the effect of noisy and missed detections of OpenPose, we use an iterative back-propagation based method using inverse kinematics to lift key points from $2D$ to $3D$ as suggested by~\cite{9093516}. 

\subsubsection{Speech Encoder}
Our network starts with a speech encoder $SE$ used to learn contextual embeddings from the input speech segment. We use melspectrograms representation of speech $S$ with $80$ mel frequency bins. We project the input melspectrogram of shape $T_s\times80$ where $T_s$ is the number of STFT time steps into a dense continuous space using a linear layer. Following this, a standard positional encoding is applied similar to~\cite{10.5555/3295222.3295349}. We then use $N$ multi-head self-attention layers with $M$ heads and also a position-wise feed-forward layer. Each of the two sub-layers has residual connections and layer normalization. Our speech encoder produces outputs of dimension $d_{model}$ and the final embedding is represented by $Z=\{z_1, z_2,..z_n\}$. 

\subsubsection{Pose Decoder}
The task of the pose decoder $PD$ is to attend to the learned context from the speech encoder and the previous time steps to generate pose at the current time step. Input to pose decoder is the masked pose sequence of dimension $m \times 151 $. Pose $p_i$ for every $ith$ frame is initially projected to a dense continuous space using a linear layer. To guide the generation process without the special symbol denoting the start and end of the sequence as done while handling discrete vocabulary, we add a counter in the range of $0-1$ to every frame's pose similar to~\cite{10.1007/978-3-030-58621-8_40}. We then use positional encoding and $N$ masked multi-head self-attention layers with $M$ heads which produce embeddings of dimension $d_{model}$. This is followed by $N$ multi-head cross-modal attention layers attending to speech embeddings modeling the translation between speech and pose. Finally, a position-wise feed-forward layer followed by a dense layer is used to generate the predicted pose $\hat{p_i}$ for $ith$ frame given as,
\begin{equation}
  \hat{p_{i}} = PD(\tilde{p}_{i-1} \vert z_{1:n}, \tilde{p}_{1:i-2})
\end{equation}
Here, $z_{1:n}$ represents the contextual embedding of the speech input segment $S$ obtained from the speech encoder, $\tilde{p}_{i-1}$ is the pose embedding for pose frame $i-1$. The predicted pose is used to calculate the regression loss for the network given as,
\begin{equation}
    \mathcal{L}_{Reg} = \frac{1}{m}\sum_{i=1}^{m}(p_i - \hat{p}_i)^2  
    \label{eq1}
    \vspace{-5pt}
\end{equation}

\subsubsection{Text Decoder}
The text decoder $TD$ is similar to a classical transformer decoder block for discrete vocabulary. The input to this block is the masked text translations of length $Q$. It uses positionally encoded input tokens at time step $i$ to decode the subsequent tokens in the sequence until an end-of-sequence token is produced. It also consists of masked multi-head self-attention blocks followed by cross attention with the speech context embeddings and then non-linear point-wise feed-forward layers. 
\begin{equation}
 {h_{i}} = TD(\tilde{t}_{i-1} \vert z_{1:n}, \tilde{t}_{1:i-2})
 \vspace{-5pt}
\end{equation}
where ${h_{i}}$ is the output embedding from the text decoder, $\tilde{t}_{i-1}$ represents the word embedding for spoken language token $t_{i-1}$ at $(i-1)th$ time step and $z_{1:n}$ represents the contextual embedding of the speech input $S$. The output at every time step ${h_{i}}$ is then used to get a softmax probability over the target vocabulary $V$. A cross entropy loss given in equation~\ref{eq2} is computed and back-propagated over the network.
\begin{equation}
    \mathcal{L}_{Xent} = 1 -\prod_{i=1}^{Q}\sum_{j=1}^{V}p(t_i^j)p(\hat{t}_i^j\vert  {h_{i}})
    \label{eq2}
\end{equation}
Here, $Q$ is the sentence length, the $V$ is the target vocabulary, $p(t_i^j)$ is the ground truth probability and $p(\hat{t}_i^j\vert  {h_{i}})$ is the predicted probability for the $ith$ token to be $t^j$ .

\subsubsection{Cross-Modal Discriminator}
\begin{figure}[t]
  \centering
  \includegraphics[width=\linewidth]{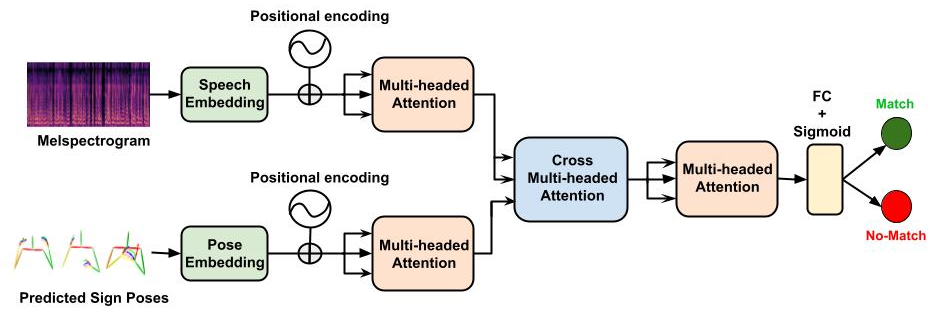}
  \caption{We fuse the two modalities with a cross-modal discriminator, allowing the network to correlate between poses and speech time-steps. This fused representation is used to perform a binary classification to classify ``match" or ``no-match".}
  \label{fig:speech_production}
  \vspace{-10pt}
\end{figure}

We introduce a matching network which is used as a cross-modal $CM$ discriminator. This network is used to match the speech segments and the corresponding sign pose sequences. Figure \ref{fig:speech_production} shows the design of our cross-modal matching network. 
It consists of separate speech and pose embedding layers which first learn a high dimensional embedding of input speech and pose sequence. We have multiple self attention blocks to learn attention aware representations for each modality which are then fused using a multi-headed cross attention block.
The cross attention fusion of two modalities learns to embed speech encoding into the pose space and learns the relationship between them. The cross-modal attention block is followed by a position-wise feed-forward layer and a few self-attention blocks. Finally, a fully-connected layer with sigmoid non-linearity is used to obtain a probability signifying ``match" or ``no-match".
In our experiments, we show the effectiveness of cross-modal discriminator in improving the network's performance. 

\begin{table}[t]
  \setlength{\tabcolsep}{1.2pt}
\footnotesize
  \caption{We report scores from various baseline approaches along with our proposed network. We can see that our network outperforms other baselines in the text to sign language and speech to sign language settings. Since ~\cite{surrey858417} was trained on Phoenix 14T~\cite{8578910} dataset in the original paper, we also report the scores on the same data as a reference (last row).}
  \label{tab:text_speech_to_SL}
  \centering
  \begin{tabular}{ c c c c c }
    \toprule
    \multicolumn{1}{c}{\textbf{Dataset}} &
    \multicolumn{1}{c}{\textbf{Approach}} &
    \multicolumn{1}{c}{\textbf{Task}} &
    \multicolumn{1}{c}{\textbf{DTW}$\downarrow$} & 
    \multicolumn{1}{c}{\textbf{PCK}$\uparrow$} \\
    \midrule

    ISL dataset  &Saunders et al.~\cite{surrey858417} & Text to SL & 16.26 & 40.14\\
    ISL dataset & LSTM + attention & Speech to SL & 17.37 & 39.30 \\
    ISL dataset & Ours w/o multi-tasking & Speech to SL & 16.71 & 41.60\\
    ISL dataset & Ours w/o discriminator & Speech to SL &14.94 & 49.40\\
    \textbf{ISL dataset} & \textbf{Ours} & \textbf{Speech to SL} & \textbf{14.05} & \textbf{53.30}\\
    \midrule
    Phoenix 14T~\cite{8578910} & Saunders et al.~\cite{surrey858417}  & Text to SL & $11.44$ & $38.12$    \\
    \bottomrule
  \end{tabular}
  \vspace{-20pt}
\end{table}

\subsubsection{Joint Training of Transformer with Cross-Modal Discriminator}
We train the multi-task transformer with cross-modal discriminator jointly in a conditional generative adversarial network~\cite{mirza2014conditional} setting to produce more realistic pose sequences. Our proposed multi-task transformer acts as a generator $G$, and the cross-modal matching network acts as a discriminator $D$, and they compete in a min-max game given by the objective function,

\vspace{-15pt}
\begin{multline}
  \min_{G} \max_{D} \mathcal{L}_{GAN}(G,D) =  \mathbb{E}[\log D(X \vert Y)]\\ + \mathbb{E}[\log(1-D(G(\hat{X}\vert Y)]
  \label{eq3}
  \vspace{-10pt}
\end{multline}
where, $\hat{X}$ represents the generated pose sequence, $X$ represents the target pose sequence, and $Y$ represents the ground truth input speech. 
Thus, the overall loss for the network is given as a weighted sum of losses in Equations \ref{eq1}, \ref{eq2} and \ref{eq3}, 
\begin{equation}
    \mathcal{L}_{Total} = \lambda_{Reg} \mathcal{L}_{Reg} + \lambda_{Xent} \mathcal{L}_{Xent} + \lambda_{GAN} \mathcal{L}_{GAN}
    \vspace{-5pt}
\end{equation}
where $\lambda_{Reg}$ is the regression loss weight, $\lambda_{Xent}$ is the recognition loss weight, and $\lambda_{GAN}$ is the weight for adversarial loss. 

\subsection{Implementation details}
We have implemented all our models using the PyTorch framework~\cite{pytorch}. In all the transformer layers, we use an embedding size of $d_{model}=512$, $N=2$ layers and number of heads $M=8$. We use Xavier initialization and Adam optimizer with an initial learning rate of $10^{-3}$ for training the transformer and the cross-modal discriminator. We also use data augmentation like predicting multiple-frame poses at each time step as done by~\cite{10.1007/978-3-030-58621-8_40}. We predict $10$ frames at every time step to penalize the network heavily for producing mean poses. We have set $\lambda_{Reg}=1, \lambda_{GAN}=10^{-4}, \lambda_{Xent}=10^{-3}$ in our experiments.

\section{Results and Evaluation}
We evaluate the quality of the generated sign language pose sequences from different models using Dynamic Time Warping (DTW) and Probability of Correct Keypoints (PCK) scores. 
DTW finds an optimal alignment between two time series by non-linearly warping them. Lower DTW corresponds to better pose generations. On the other hand, PCK is used in several pose detection and generation works~\cite{6380498},~\cite{8954219}. It evaluates the probability of pose key points to be close to the ground truth key points up to a threshold of $\alpha = 0.2$. Higher PCK corresponds to better pose generations.

\begin{figure}[t]
  \centering
  \includegraphics[width=\linewidth]{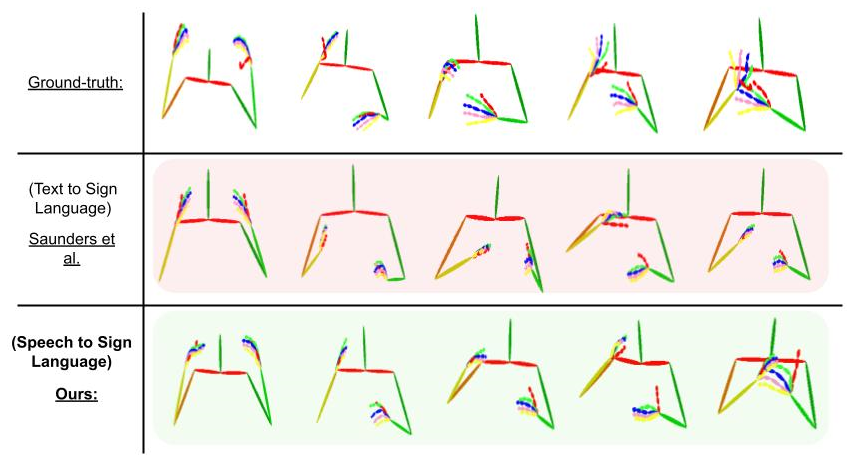}
  \caption{We show a graphical comparison between the generated outputs from the state-of-the-art text to sign language and our speech to sign language models. We can observe that the poses generated by our model are closer to the ground-truth pose sequence indicating its superiority.}
  \label{fig:comp_disc}
  \vspace{-10pt}
\end{figure}

We start by comparing with the state-of-the-art text to sign language model~\cite{surrey858417} trained on the proposed ISL dataset as shown graphically in Figure~\ref{fig:comp_disc}.
Since ours is the first work to deal with speech to sign language generation, we do not have directly comparable networks for the given task. Thus, we also establish several baselines for the first time, as shown in  Table~\ref{tab:text_speech_to_SL}. Our first proposed baseline is a modified speech-to-text~\cite{7472621} network consisting of an LSTM seq2seq network with an attention mechanism. We modified the network for the regression task of generating pose key points instead of character probabilities. This network serves as a comparison between the RNN based approach and our transformer-based method. We observe that using an LSTM-based network leads to mean pose being generated at each time step, severely impacting the outcome. We also train a single task network by removing the text decoder block from our architecture. As evident from Table~\ref{tab:text_speech_to_SL}, incorporating an auxiliary task of generating text sequences from speech helps in the main task of sign language generation. Finally, we also evaluate the importance of the cross-modal discriminator by training a model without the discriminator. As seen from Table~\ref{tab:text_speech_to_SL}, our network with cross-modal discriminator achieves significantly better metric scores (DTW:$14.05$, PCK:$53.33$) compared to the best available text to sign language model~\cite{surrey858417} and other speech to sign language baselines. Since~\cite{surrey858417} was originally trained on the Phoenix 14T~\cite{8578910} dataset, we also report the reference scores for the reader in the same table.

To highlight the importance of the cross-modal matching network, we perform an ablation experiment with other discriminators. We use a $1$D convolutional discriminator to classify generated pose sequence as real or fake, similar to the one proposed by Saunder et al. in~\cite{surrey858417}. 
As seen in Table~\ref{tab:compare_D}, other types of discriminators are detrimental for the network and work poorer than the ``without" discriminator setup.

\begin{table}[t]
\footnotesize
  \caption{Quantitative evaluation for speech to sign language generation using different discriminators.}
  \label{tab:compare_D}
  \centering
  \begin{tabular}{ c c c }
    \toprule
    \multicolumn{1}{c}{\textbf{Approach}} &
    \multicolumn{1}{c}{\textbf{DTW}$\downarrow$} &
    \multicolumn{1}{c}{\textbf{PCK}$\uparrow$} \\
    \midrule
    Ours w/o discriminator  & 14.94 & 49.40 \\
    Ours with conv. disc & 15.07& 45.41\\
    Ours with pose-only disc & 15.34  & 43.76\\
    \textbf{Ours}  & \textbf{14.05} & \textbf{53.33}\\
    \bottomrule
  \end{tabular}
  \vspace{-15pt}
\end{table}

\section{Conclusion}
In this work, we introduce the task of translating the spoken language to sign language pose sequences to make way for two-sided communication between the hearing impaired and the rest of the world. We propose a new ISL dataset with speech modality and avoid the expensive gloss annotations. We are the first to achieve state-of-the-art results for the task of speech to sign language generation using a multi-task transformer network coupled with a cross-modal discriminator. Our work paves the way for future research in speech to sign language generation that may one day result in real-time sign language translation. 




\bibliographystyle{IEEEtran}

\bibliography{mybib}

\end{document}